\documentclass[conference]{IEEEtran}
\IEEEoverridecommandlockouts
\usepackage{cite}
\usepackage{amsmath,amssymb,amsfonts}
\usepackage{graphicx}
\usepackage{textcomp}

\usepackage{mathptmx}

\usepackage[T1]{fontenc}
\usepackage[latin9]{inputenc}
\usepackage{geometry}
\usepackage{textcomp}

\usepackage{algorithm}
\usepackage{algpseudocode}
\usepackage{amsthm}
\usepackage{amsmath}
\usepackage{romannum}
\usepackage{verbatim}
\usepackage{notoccite}
\usepackage{hyperref}

\hypersetup{colorlinks=true, linkcolor=blue, citecolor=blue,}

\def\BibTeX{{\rm B\kern-.05em{\sc i\kern-.025em b}\kern-.08em
    T\kern-.1667em\lower.7ex\hbox{E}\kern-.125emX}}
\begin{document}

\title{Multiobjective Simulated Annealing Algorithm for Minimum Weight Minimum Connected Dominating Set Problem\\}

\author{\IEEEauthorblockN{1\textsuperscript{st} Hayet Dahmri}
\IEEEauthorblockA{\textit{Department of Computer Science} \\ \textit{Mechatronics Laboratory (LMETR)-E1764200} \\
\textit{University of Ferhat Abbas - S\'{e}tif 1}\\
S\'{e}tif, Algeria \\
hayet.dahmri@univ-setif.dz}
\and
\IEEEauthorblockN{2\textsuperscript{nd} Salim Bouamama}
\IEEEauthorblockA{\textit{Department of Computer Science} \\ \textit{Mechatronics Laboratory (LMETR)-E1764200} \\
\textit{University of Ferhat Abbas - S\'{e}tif 1}\\
S\'{e}tif, Algeria\\
salim.bouamama@univ-setif.dz}
}
\maketitle

\begin{abstract}
Minimum connected dominating set problem is an NP-hard combinatorial optimization problem in graph theory. Finding connected dominating set is of high interest in various domains such as wireless sensor networks, optical networks and systems biology. Its weighted variant named minimum weight connected dominating set is also useful in such applications. In this paper, we propose a simulated annealing algorithm based on a greedy heuristic for tackling a variant of the minimum connected dominating set problem and that by exploiting two objectives together namely the cardinality and the total weight of the connected dominating set. 
Experimental results compared to those obtained by a recent proposed research show the superiority of our approach.\\

 \begin{small}
 \textit{\textbf{keywords}--Minimum connected dominating set, Minimum weight connected dominating set, Simulated annealing, Greedy heuristic, Wireless sensor network.}
 
 \end{small}
 
\end{abstract}

\section{Introduction}
\label{sec1}
The minimum connected dominating set  problem (MCDS) and its variants are among  the well-known combinatorial optimization problems in graph theory with applications to many fields especially in wireless network communications \cite{wang2005distributed:mwcds, zhang2008minimum:mwcds, misra2009minimum:ASNET, shukla2013construction:wsn, chinnasamy2019minimum:VANET}, optical networks \cite{sen2010brief:optic, sen2008sparse:optic} and  systems biology \cite{milenkovic2011dominating}.
 More details in this context can be found in \cite{blum2004connected, du2012connected}.

Given a simple undirected graph $G =(V, E)$, a dominating set $D$ is a subset of $V$, such that each vertex in $D$ is adjacent to at least one vertex from $D$ and the subgraph induced by $D$ is connected. The MCDS problem asks for a connected dominating set with minimum cardinality (size). If a positive weight is associated with each vertex of $V$, the minimum weight connected dominating set problem (MWCD) looks for a connected dominating set with minimum total weight. \\
Both MCDS and MWCDS belong to the class of NP-hard problems \cite{gary1979computers}. Due to their difficulty and the potential benefits of solving them, considerable works have been conducted in this regard. Most of them are based on metaheuristic algorithms, which are approximate approaches that can find reasonably near-optimal solution in an acceptable computation time, rather than exact algorithms
that guarantee the optimality of the returned solutions but in an exponential time.\\
As example of such approaches, Morgan and Grout \cite{morgan2007metaheuristics} introduced the first metaheuristic to deal with MCDS problem. The latter combines tabu search and simulated annealing algorithm. Jovanovic et \textit{al.}~\cite{jovanovic2013ant} proposed an ant colony optimization algorithm (ACO) with greedy heuristics. Recently, Li et \textit{al.}~\cite{li2017grasp} presented a greedy randomized adaptive search procedure (GRASP) that incorporates a tabu search as a local enhancement process. Besides, Wu et \textit{al.}~\cite{wu2017restricted} developed a tabu search procedure (RSN-TS) based on a restricted swap-based neighborhood. The authors conducted a considerable number of experimental tests to show that RSN-TS outperforms GRASP and ACO both in terms of solution quality and computation time. Later, Hedar et \textit{al.}~\cite{hedar2019two} implemented two methods for solving MCDS problem. The first one is a memetic algorithm and the second one is a simulated annealing. The performance of both approaches when applied to MCDS problem on common benchmark instances is better than ACO and GRASP but less than RSN-TS based on results reported in literature. However, few researches have been carried out for MWCDS problem. A hybrid genetic algorithm (HGA) and a population-based iterated greedy (PBIG) algorithm were proposed in \cite{dagdeviren2017two}. More recently Bouamama et \textit{al.}~\cite{bouamama2019algorithm} developed a hybrid ant colony optimization approach combined with a reduced variable neighborhood search (ACO-RVNS) to solve  both MCDS and MWCDS. In this algorithm, MCDS is considered as MWCDS with unit weight of one to every vertex of the input graphs. It was shown that ACO-RVNS outperforms RSN-TS, PBIG and HGA on all available benchmark sets especially for large problem instances.\\
\indent One should mention that all previous approaches have something in common: they optimize only a single objective function such as minimizing the size of the connected dominating set (CDS) and minimizing its total weight for MCDS and MWCDS, respectively. To the best of our knowledge, the only existing approach in the literature that considered these two objectives together was presented in \cite{rengaswamy2017multiobjective}.~The authors of this study firstly defined the minimum weight minimum connected dominating set problem (MWMCDS) of which the aim is to minimize simultaneously the size and the total weight of generated CDS. Then, they proposed a multiobjective genetic algorithm (MOGA) based on scalarization model to deal with MWMCDS problem.\\
\indent In this work, we propose to solve the MWMCDS problem with a so-called multiobjective greedy simulated annealing (GSA) algorithm. The concept of scalarization is used to provide a trade-off between size and total weight and a greedy heuristic is needed to seed simulated annealing with good initial solutions and generating neighbors.\\
\indent The remainder of the paper is organized as follow: Section \ref{sec2} describes the problem. Section \ref{sec3} discusses the proposed GSA algorithm. The experimental results are given in Section \ref{sec4}. Section \ref{sec5} concludes the paper.

\section{Problem statement}
\label{sec2}
\begin{figure}[h]
\centering
\includegraphics[scale=0.4]{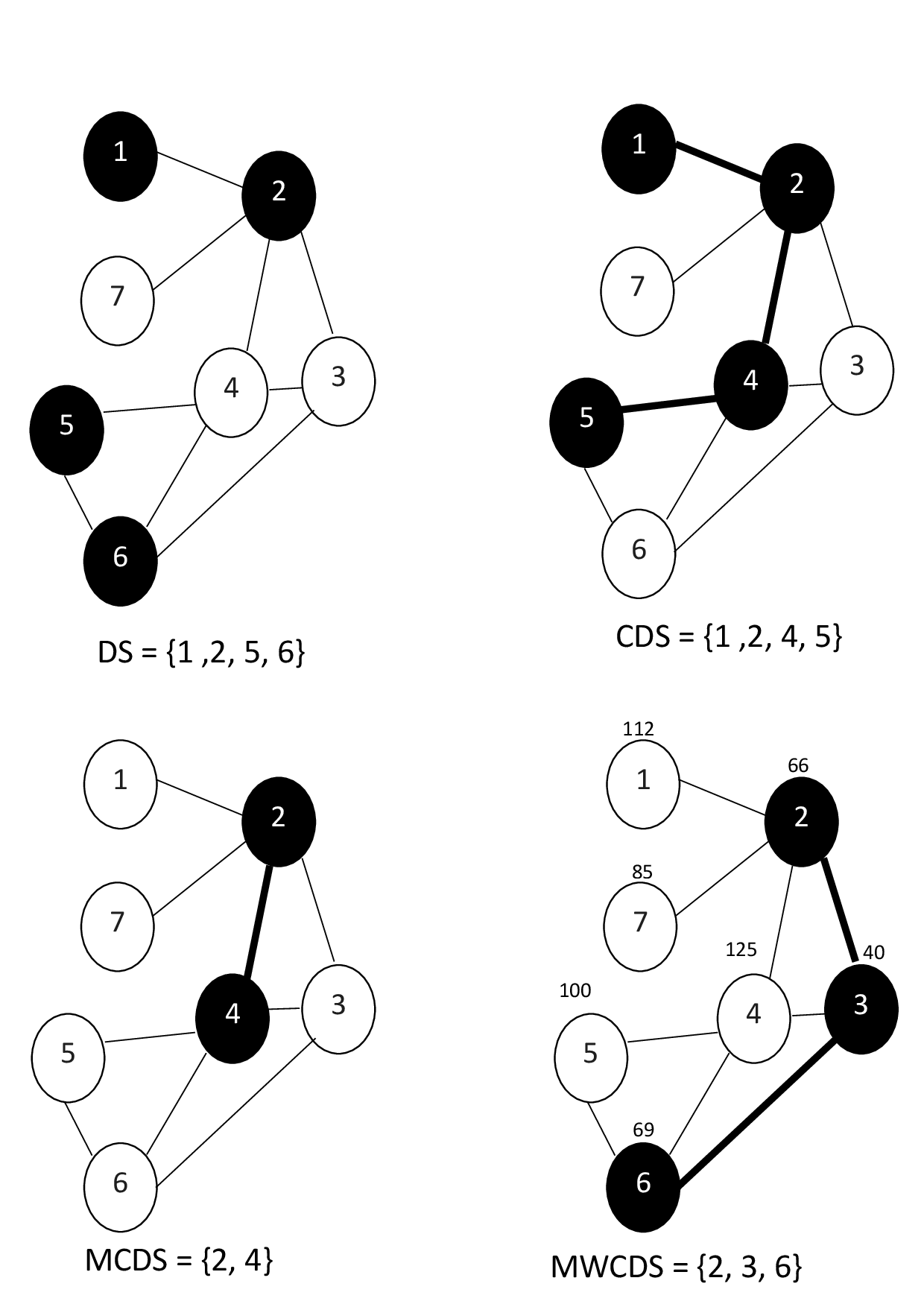}
\caption{Example of DS (Dominating Set), CDS (Connected DS),  MCDS (Minimum CDS) and MWCDS (Minimum weighted CDS) in a simple undirected graph.}
\label{fig:myFig1}
\end{figure}

Given a simple undirected weighted graph $G =(V,E)$ where $V =\{1, 2, \cdots, n \}$ represents the set of vertices, and $E \subset  V \times V$ represents the set of edges. Two vertices are said to be adjacent or neighbors if they are joined by an edge. The set of neighbors of $v$ is denoted by $N(v) = \{u \in V \mid (v,u) \in E \}$. A subset $D\subseteq V$ is called a dominating set if each vertex $v \in V$ is either in $D$ or adjacent to at least one vertex in $D$.  If $D$ is a dominating set and its induced subgraph $G(D)$ is connected, then $D$ is called a connected dominating set. Vertices in the dominating set are called the dominators. Fig. \ref{fig:myFig1} gives an illustrative example explaining these definitions on a simple undirected graph with 7 vertices and 9 edges. Dominators are highlighted with a black background and white lettering. An MWMCDS problem consists of a simple undirected connected graph $G = (V, E,w)$ where $w : E \mapsto R^{+}$ is a weight function that assigns a positive weight value $w_{(v,u)}$ to each edge $ (v,u) \in E $ of the graph. This problem can thus be expressed as

\begin{tabular}{ll}
 &					\\	
 \textbf{minimize} & $ \{ F_{c}(D), F_w(D) \} $\\
 &					\\							
 \textbf{subject to} & $\forall\ v \in V\backslash D: N(v)\cap D \neq \emptyset$,\\	
 & $D \subseteq V$, \\	
 & $G(D)$ is connected. \\	
 &					\\	 
\end{tabular}

\noindent In the above definition, we look for a connected dominating set $D \subseteq V$ (a candidate solution) in which two objective functions are simultaneously minimized. Let $\vert D \vert$ represents the cardinally of $D$. The first objective function $ F_{c}(D) := \vert D \vert $, named as the cardinality objective function, intends to minimize the size of the candidate solution while the second objective function $F_{w}(D)$ named as the weight objective function, intends to minimize its total weight. $F_{w}(D)$ is calculated as follow. 

\begin{align}
& F_{w}(D) := F_{w1} + F_{w2}   \\
& F_{w1} = \sum\limits_{ ((u,v) \in E) \wedge (u \in D \wedge v \in D)} {w_{(u,v)}}  \\
& F_{w2} = \sum\limits_{ (u\in V\setminus D)}  \textbf{min} \{ w_{(u,v)} \mid (u,v) \in E \wedge v \in D\} 
\end{align}   

\section{Multiobjective Greedy Simulated Annealing algorithm GSA for MWMCDS problem}
\label{sec3}

In the following we describe the main components of GSA algorithm that we developed with the aim of obtaining high quality solutions to the MWMCDS problem. As mentioned before, GSA is a multiobjective greedy simulated annealing algorithm  based on scalarization. 

\subsection{Objective function}
Given a candidate solution $S$, with respect to the objective functions $F_{c}(D)$ and $F_{w}(D)$ defined in Section \ref{sec2}, the MWMCDS aggregated objective function can be defined as:
\begin{equation}
    F  :=  \alpha \times F^{'}_{c}(D) +  \beta \times F^{'}_{w}(D)
\label{eq3}
\end{equation} 

Where  $ \alpha \in [0,1]$ and  $ \beta \in [0,1]$ are two  algorithm-based parameters such that $\alpha + \beta =1$ and $\alpha = \beta$. In addition, $F^{'}_{c}(\cdot)$ and $F^{'}_{w}(\cdot)$  are the normalized values of $F_{c}(\cdot)$ and  $F_{w}(\cdot)$, based on the total number of vertices and total weight in the network respectively.

\subsection{Greedy heuristic}
\label{gr}
A feasible solution is greedily constructed as follow. Given an empty solution $S$. All vertices in $V$ are colored initially with color WHITE.  Let $d_{S}(v)$ denotes the current degree of vertex $v$  which represents the number of WHITE neighbors of $v$ with respect to $S$. Firstly, the color of the vertex with highest current degree becomes GRAY. Then, we repeatedly select the GRAY vertex with the maximum value of $d_{S}(\cdot) $ to be included in $S$. Besides, its color becomes BLACK while the color of their WHITE neighbors will be changed to GRAY. The process is continued until all vertices are either colored BLACK or GRAY (no WHITE vertices are left).\\
The first vertex $v^{first}$ is chosen as follow:
\begin{equation}
 v^{first} \leftarrow  \textbf{argmax}  \{  d_{S}(v)  \mid  v\in V\}. 
\end{equation} 
The remaining vertices to be placed in $S$ are chosen as follow:
 \begin{equation}
 v^{*} \leftarrow  \textbf{argmax}  \{ d_{S}(v)  \mid  (v\in V \setminus S) \; \wedge \; (\textnormal{color}(v)= \textbf{GRAY}) \}.
\end{equation}

\subsection{Representation}
A problem instance is mapped to an edge weighted graph $G(V, E, w)$ where each vertex $v$ in $V$ is represented by a unique integer number from $ \{0,1,\cdots,n-1\} $, where $n$ denotes the size of $V$, that is, $n =\vert V \vert$. A candidate solution $S$ is coded by a vector of fixed length $n$ of which each element can take values 0 or 1 depending on if the corresponding vertex belongs to $S$ or not.

\subsection{Multiobjective simulated annealing framework}
Simulated Annealing (SA) \cite{kirkpatrick1983optimization}  is a well known metaheuristic approach that has been applied successfully to a large  number of combinatorial optimisation problems. SA  is both a single-solution based algorithm and exploitation oriented. Our algorithm named GSA is a standard SA algorithm improved by 
starting with a good initial solution based on the greedy heuristic previously defined (see Section \ref{gr}) and the neighbors are also generated either greedily or randomly. Moreover, GSA uses a useful calculation method for changing temperature. These improvments ensure the computational efficiency and improve the quality
of obtained solutions. A high level description of GSA algorithm is illustrated in Alg. \ref{alg1}. The \textsf{generate\_initial\_solution ($sol\_size$)} procedure (see Alg. \ref{alg2}) generates $sol\_size$ solutions for the initial solution pool. Out of these solutions, one is generated greedily by applying \textsf{generate\_greedy()} procedure (see Alg. \ref{alg3}) while the others are generated randomly using \textsf{generate\_random()} procedure (see Alg. \ref{alg4}).\\
On line 17 of Alg. \ref{alg1} - \textsf{accept($Temperature,S,S'$)} -, if the incumbent solution $S$ is worst than its neighbor $S'$ , then $S'$ will be kept, otherwise it may be accepted or rejected depending one the Metropolis condition.
On line 20 - \textsf{change\_temperature($Temperature,k$)} - after $k$ (here $k$ is set to 3) consecutive iterations at the same temperature, the temperature will be decreased by a factor of 1-$\gamma$ ($\gamma$ = 0.9). If the temperature takes value lower than 1, we use a temperature reheating and it returns to the initial value $T_0$.\\
The neighbor of a candidate solution is obtained either greedily using procedure \textsf{neighbor\_greedy()} (see Alg. \ref{alg5}) or randomly using procedure \textsf{neighbor\_random()} with respect to a probability distribution $p$. In this context, both procedures follow the same basic steps except that the last chooses the input solution and the dominator vertices randomly.

\begin{algorithm}[!h]
\caption{GSA for the MWMCDS problem}
\begin{algorithmic}[1]
\State \textbf{input}: $A~ problem ~instance~ (G,V,E,w),~ and~ parameters$ $ ~\textit{sol\_size} ~,k~ and ~T_0$
\State $\varphi \gets \ \textsf{generate\_initial\_solution(sol\_size)}$  //see Alg. \ref{alg2}
\State $Temperature \gets \ T_0$ 
\State  $S^{best} \gets argmin \{F(S)|~S \in \varphi\}~$ //see (\ref{eq3}) for the definition of F. 
\State  $S \gets S^{best}$  
\While{termination condition not satisfied}
\State $p \gets random~number~uniformly~distributed~over~[0,1]$                         
\If {$p>0.5$} 
   \State $S' \gets \textsf{neighbor\_greedy(S)}$ //see Alg. \ref{alg5}
\Else
   \State $S  \gets pick~ a~random ~solution ~from ~\varphi $   
   \State $S' \gets \textsf{neighbor\_random(S)}$         
\EndIf
\If {$F(S') < F(S^{best})$} 
\State $S^{best} \gets S'$
\EndIf
\If {\textsf{accept(Temperature,S,S')}}
 \State $ S  \gets S'$
\EndIf
 \State $Temperature \gets \textsf{change\_temperature(Temperature,k)}$ 
\EndWhile    
\State \textbf{output}: $\{ S^{best}, F_c(S^{best}),F_w(S^{best}) \}$
\end{algorithmic}
\label{alg1}
\end{algorithm}

\begin{algorithm}[!h]
\caption{generate\_initial\_solution (sol\_size)}
\begin{algorithmic}[1]
\State \textbf{input}: $sol\_size$
\State $\varphi \gets \emptyset$ 
\State  $S \gets \textsf{generate\_greedy()}$  //see Alg. \ref{alg3}
\State  $\varphi \gets \varphi  \cup \{S\}$
\For {$i \gets ~ 2 ~ to ~ sol\_size$ }
\State   $S \gets \textsf{generate\_random()}$  //see Alg. \ref{alg4}
\State   $\varphi \gets \varphi \cup \{S\}$
\EndFor
\State  \textbf{output}: $\varphi=\{S_1, S_2,\dots, \textrm{S}_\textrm{sol\_size}\}$
\end{algorithmic}
\label{alg2}
\end{algorithm}

\begin{algorithm}[!h]
\caption{generate\_greedy()}
\begin{algorithmic}[1]
\State $S \gets \emptyset$ 
\State  $v^{first} \gets~ \textbf{argmax}\{d_s(v)~ |~ v \in V\}$  
\State $v^{first}.color \gets GRAY$
\ForAll {$v \in V \setminus \{v^{first}\}~$} 
\State $v.color \gets WHITE$       
\EndFor
\Repeat
\State $ v^{\ast} \gets ~ \textbf{argmax}\{d_s(v)~ |~ (v \in V \setminus S)~ \land~ (color(v)=GRAY)\}$ 
 \State $v^{\ast}.color \gets BLACK$
 \State  $S \gets S  \cup \{v^{\ast}\}$
\ForAll { $ v \in N(v^{\ast}) \setminus S$ }           
 \State $v.color \gets GRAY$         
\EndFor    
\Until (All vertices are colored either BLACK or GRAY)
\State  \textbf{output}:$~S$
\end{algorithmic}
\label{alg3}
\end{algorithm}

\begin{algorithm}[!h]
\caption{generate\_random()}
\begin{algorithmic}[1]
\State $S \gets \emptyset$ 
\State  $v^{first} \gets a ~random ~vertex~ from~ V$  
\State $v^{first}.color \gets GRAY$
\ForAll {$v \in V \setminus \{v^{first}\}~$} 
\State $v.color \gets WHITE$       
\EndFor
\Repeat 
\State $ v^{\ast} \gets a ~random ~vertex ~from~ GRAY~ vertices$ 
 \State $v^{\ast}.color \gets BLACK$
 \State  $S \gets S  \cup \{v^{\ast}\}$
\ForAll { $ v \in N(v^{\ast}) \setminus S$ }           
 \State $v.color \gets GRAY$         
\EndFor    
\Until (All vertices are colored either BLACK or GRAY)
\State  \textbf{output}:$~S$
\end{algorithmic}
\label{alg4}
\end{algorithm}

\begin{algorithm}[!h]
\caption{Procedure neighbor\_greedy($S$)}
\begin{algorithmic}[1]
\State \textbf{input}: $an~ incumbent~ solution ~S$
\State  $S' \gets S$
\State  $v \gets \textbf{argmin}\{|N(v)|, v \in S' \}$ 
\State $v.color \gets GRAY$
\State $S' \gets S' \setminus \{v\}$
\ForAll {$vertex ~v^p \in N(v)$}
\If{$(v^p.color == GRAY) \land (N(v^p) \cap (S' \setminus \{v\})=\emptyset)$}
\State $v^p.color \gets WHITE$
\EndIf
\EndFor
\Repeat 
\State $ v^{\ast} \gets ~ \textbf{argmax}\{\textrm{d}_\textrm{s'}(v)~ |~ (v \in V \setminus S') \land (color(v)=GRAY)\}$ 
 \State $v^{\ast}.color \gets BLACK$
 \State  $S' \gets S' \cup \{v^{\ast}\}$
\ForAll { $ v \in N(v^{\ast}) \setminus S'$ }           
 \State $v.color \gets GRAY$         
\EndFor
\Until (All vertices are colored either BLACK or GRAY)
\State  \textbf{output}: $S'$
\end{algorithmic}
\label{alg5}
\end{algorithm}

\section{EXPERIMENTAL EVALUATION}
\label{sec4}
 The proposed algorithm GSA was implemented using C++ language. The experimental results were obtained on a PC with an  Intel Core i5-1135G7 2.40GHz processor and 8.0 GB of memory.\\
Its performance was compared against two recent algorithms from the literature, namely MOGA \cite{rengaswamy2017multiobjective} and mcds \cite{nazarieh2016identification}. The results of the two approaches are reproduced from \cite{rengaswamy2017multiobjective}. GSA was evaluated on the same benchmark set introduced in \cite{rengaswamy2017multiobjective} where each instance consists of a simple undirected edge-weighted graph modeling a data transfer system where every vertex transfers data at instant $t$ with probability $P_t$ and this data can be dropped with probability $P_d$.
The distance traveled by the data in the transfer is represented indirectly by the weight, which expressed by the energy consumed during the travel. 
Thus, the energy consumed in case of successful transfer is equal to the distance traveled by the data, and in the case of failed transfer is equal to half of distance traveled. The energy consumed in transfer of all data in the network is represented by energy consumption.\\ 

\begin{figure}[htbp]
\centering
\includegraphics[width=3.3in, height=3.5in]{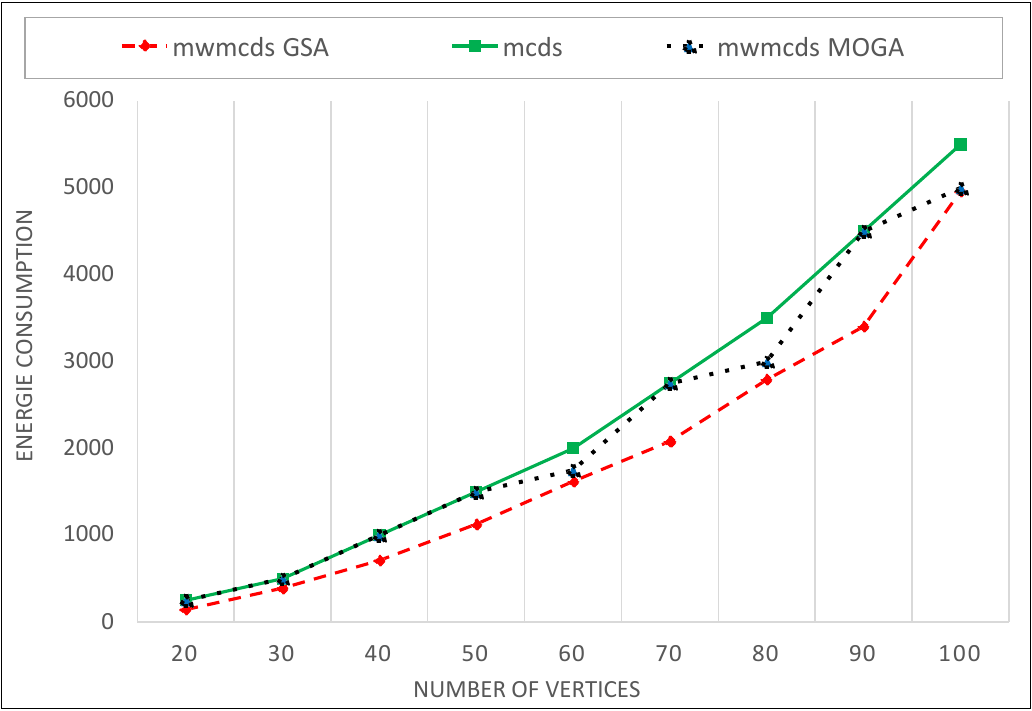}
\caption{Energy consumption in GSA, MOGA and mcds}
\label{fig1}
\end{figure}

The performance of GSA against MOGA and mcds with respect to energy consumption for 100 instances of data transfer is shown in Fig. \ref{fig1}. It can be seen that GSA consumes the least amount of energy in all cases compared with mcds. Compared to MOGA, the energy requirements of GSA are lesser in all cases except the scenario ($n=100$) where both algorithms give similar results.

\begin{figure}[htbp]
\centering
\includegraphics[width=3.3in, height=3.5in]{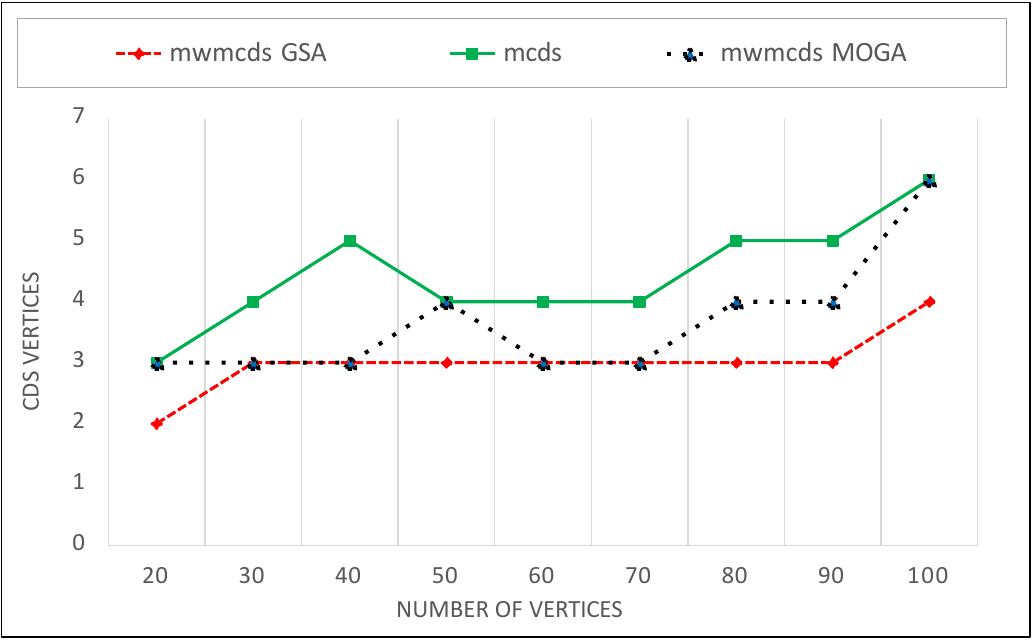}
\caption{Size of CDS in GSA, MOGA and mcds}
\label{fig2}
\end{figure}

Fig. \ref{fig2} represents the number of dominators vertices produced by GSA, MOGA and mcds for different networks.  Here too it can be seen that GSA performs better than mcds in all cases by producing CDS of minimal size. Compared to MOGA, GSA give the same results in 4 cases (n= 30, n=40, n=60 and n=70), in all other cases GSA performs the best.

From previous results, it can be deduced that the proposed algorithm GSA generates more useful solutions for the MWMCDS problem than mcds and MOGA.

\section{Conclusion}
\label{sec5}
In this work we have proposed a new approach called GSA for tackling the minimum weight minimum connected dominating set problem in wireless sensor networks. Two objectives
are involved together, namely the weight and the size of the connected dominating set. GSA integrates a simulated annealing algorithm, which is seeded by a greedy constructive
heuristic and employed in two hybridization models.
The performance of the proposed algorithm was assessed and compared against a recent multiobjective genetic algorithm. For future work, it would be interesting to use Pareto
optimization techniques instead of scalarization that reduces a multiobjective optimization problem to single objective optimization problem by a linear combination.

\bibliographystyle{IEEEtran}
\bibliography{ref}

\begin{thebibliography}{10}
\providecommand{\url}[1]{#1}
\csname url@samestyle\endcsname
\providecommand{\newblock}{\relax}
\providecommand{\bibinfo}[2]{#2}
\providecommand{\BIBentrySTDinterwordspacing}{\spaceskip=0pt\relax}
\providecommand{\BIBentryALTinterwordstretchfactor}{4}
\providecommand{\BIBentryALTinterwordspacing}{\spaceskip=\fontdimen2\font plus
\BIBentryALTinterwordstretchfactor\fontdimen3\font minus
  \fontdimen4\font\relax}
\providecommand{\BIBforeignlanguage}[2]{{%
\expandafter\ifx\csname l@#1\endcsname\relax
\typeout{** WARNING: IEEEtran.bst: No hyphenation pattern has been}%
\typeout{** loaded for the language `#1'. Using the pattern for}%
\typeout{** the default language instead.}%
\else
\language=\csname l@#1\endcsname
\fi
#2}}
\providecommand{\BIBdecl}{\relax}
\BIBdecl

\bibitem{wang2005distributed:mwcds}
Y.~Wang, W.~Wang, and X.-Y. Li, ``Distributed low-cost backbone formation for
  wireless ad hoc networks,'' in \emph{Proceedings of the 6th ACM international
  symposium on Mobile ad hoc networking and computing}, 2005, pp. 2--13.

\bibitem{zhang2008minimum:mwcds}
J.~Zhang and C.-F. Jia, ``Minimum connected dominating set algorithm with
  weight in wireless sensor networks,'' in \emph{2008 4th International
  Conference on Wireless Communications, Networking and Mobile
  Computing}.\hskip 1em plus 0.5em minus 0.4em\relax IEEE, 2008, pp. 1--4.

\bibitem{misra2009minimum:ASNET}
R.~Misra and C.~Mandal, ``Minimum connected dominating set using a
  collaborative cover heuristic for ad hoc sensor networks,'' \emph{IEEE
  Transactions on parallel and distributed systems}, vol.~21, no.~3, pp.
  292--302, 2009.

\bibitem{shukla2013construction:wsn}
K.~K. Shukla and S.~Sah, ``Construction and maintenance of virtual backbone in
  wireless networks,'' \emph{Wireless networks}, vol.~19, no.~5, pp. 969--984,
  2013.

\bibitem{chinnasamy2019minimum:VANET}
A.~Chinnasamy, B.~Sivakumar, P.~Selvakumari, and A.~Suresh, ``Minimum connected
  dominating set based rsu allocation for smartcloud vehicles in vanet,''
  \emph{Cluster Computing}, vol.~22, no.~5, pp. 12\,795--12\,804, 2019.

\bibitem{sen2010brief:optic}
A.~Sen, S.~Banerjee, P.~Ghosh, S.~Murthy, and H.~Ngo, ``Brief announcement: On
  regenerator placement problems in optical networks,'' in \emph{Proceedings of
  the twenty-second annual ACM symposium on Parallelism in algorithms and
  architectures}, 2010, pp. 178--180.

\bibitem{sen2008sparse:optic}
A.~Sen, S.~Murthy, and S.~Bandyopadhyay, ``On sparse placement of regenerator
  nodes in translucent optical network,'' in \emph{IEEE GLOBECOM 2008-2008 IEEE
  Global Telecommunications Conference}.\hskip 1em plus 0.5em minus 0.4em\relax
  IEEE, 2008, pp. 1--6.

\bibitem{milenkovic2011dominating}
T.~Milenkovi{\'c}, V.~Memi{\v{s}}evi{\'c}, A.~Bonato, and N.~Pr{\v{z}}ulj,
  ``Dominating biological networks,'' \emph{PloS one}, vol.~6, no.~8, 2011.

\bibitem{blum2004connected}
J.~Blum, M.~Ding, A.~Thaeler, and X.~Cheng, ``Connected dominating set in
  sensor networks and manets,'' in \emph{Handbook of combinatorial
  optimization}.\hskip 1em plus 0.5em minus 0.4em\relax Springer, 2004, pp.
  329--369.

\bibitem{du2012connected}
D.-Z. Du and P.-J. Wan, \emph{Connected dominating set: theory and
  applications}.\hskip 1em plus 0.5em minus 0.4em\relax Springer Science \&
  Business Media, 2012, vol.~77.

\bibitem{gary1979computers}
M.~R. Gary and D.~S. Johnson, \emph{Computers and Intractability: A Guide to
  the Theory of NP-completeness}.\hskip 1em plus 0.5em minus 0.4em\relax WH
  Freeman and Company, New York, 1979.

\bibitem{morgan2007metaheuristics}
M.~Morgan and V.~Grout, ``Metaheuristics for wireless network optimisation,''
  in \emph{The Third Advanced International Conference on Telecommunications
  (AICT'07)}.\hskip 1em plus 0.5em minus 0.4em\relax IEEE, 2007, pp. 15--15.

\bibitem{jovanovic2013ant}
R.~Jovanovic and M.~Tuba, ``Ant colony optimization algorithm with pheromone
  correction strategy for the minimum connected dominating set problem,''
  \emph{Computer Science and Information Systems}, vol.~10, no.~1, pp.
  133--149, 2013.

\bibitem{li2017grasp}
R.~Li, S.~Hu, J.~Gao, Y.~Zhou, Y.~Wang, and M.~Yin, ``Grasp for connected
  dominating set problems,'' \emph{Neural Computing and Applications}, vol.~28,
  no.~1, pp. 1059--1067, 2017.

\bibitem{wu2017restricted}
X.~Wu, Z.~L{\"u}, and P.~Galinier, ``Restricted swap-based neighborhood search
  for the minimum connected dominating set problem,'' \emph{Networks}, vol.~69,
  no.~2, pp. 222--236, 2017.

\bibitem{hedar2019two}
A.-R. Hedar, R.~Ismail, G.~A. El-Sayed, and K.~M.~J. Khayyat, ``Two
  meta-heuristics designed to solve the minimum connected dominating set
  problem for wireless networks design and management,'' \emph{Journal of
  Network and Systems Management}, vol.~27, no.~3, pp. 647--687, 2019.

\bibitem{dagdeviren2017two}
Z.~A. Dagdeviren, D.~Aydin, and M.~Cinsdikici, ``Two population-based
  optimization algorithms for minimum weight connected dominating set
  problem,'' \emph{Applied Soft Computing}, vol.~59, pp. 644--658, 2017.

\bibitem{bouamama2019algorithm}
S.~Bouamama, C.~Blum, and J.-G. Fages, ``An algorithm based on ant colony
  optimization for the minimum connected dominating set problem,''
  \emph{Applied Soft Computing}, vol.~80, pp. 672--686, 2019.

\bibitem{rengaswamy2017multiobjective}
D.~Rengaswamy, S.~Datta, and S.~Ramalingam, ``Multiobjective genetic algorithm
  for minimum weight minimum connected dominating set,'' in \emph{International
  Conference on Intelligent Systems Design and Applications}.\hskip 1em plus
  0.5em minus 0.4em\relax Springer, 2017, pp. 558--567.

\bibitem{kirkpatrick1983optimization}
S.~Kirkpatrick, C.~D. Gelatt, and M.~P. Vecchi, ``Optimization by simulated
  annealing,'' \emph{science}, vol. 220, no. 4598, pp. 671--680, 1983.

\bibitem{nazarieh2016identification}
M.~Nazarieh, A.~Wiese, T.~Will, M.~Hamed, and V.~Helms, ``Identification of key
  player genes in gene regulatory networks,'' \emph{BMC systems biology},
  vol.~10, no.~1, p.~88, 2016.

\end{thebibliography}

\end{document}